\documentclass[journal]{IEEEtran}

\usepackage{cite}
\usepackage{hyperref}       
\usepackage{url}            
\usepackage{booktabs}       
\usepackage{amsfonts}       
\usepackage{nicefrac}       
\usepackage{microtype}      
\usepackage{amsmath}
\usepackage{amssymb}
\usepackage{amsthm}
\usepackage{bm}
\usepackage{algpseudocode}
\usepackage{algorithm}
\usepackage{mathtools}
\usepackage{subcaption}
\usepackage[pdftex]{graphicx}

\usepackage{xcolor}
\usepackage{shadethm}

\definecolor{shadethmcolor}{HTML}{F0F0F0}
\DeclareMathOperator*{\argmin}{arg\,min}

\renewcommand{\vec}[1]{\ensuremath{\mathbf{#1}}}

\newcommand{\mat}[1]{\ensuremath{\mathbf{#1}}}

\DeclareMathOperator*{\tr}{tr}
\newtheorem{theorem}{Theorem}
\newtheorem{proposition}{Proposition}

\newshadetheorem{thm}{Theorem}

\newshadetheorem{cor}[thm]{Corollary}
\newshadetheorem{prop}[thm]{Property}
\newshadetheorem{example}[thm]{Example}
\newshadetheorem{assm}[thm]{Assumption}

\newtheoremstyle{named}{}{}{\itshape}{}{\bfseries}{.}{.5em}{\thmnote{#3}}
\theoremstyle{named}
\newtheorem*{namedtheorem}{Theorem}

\begin{document}
%
\title{Learning Gaussian Graphical Models with Ordered Weighted $\ell_1$ Regularization}

\author{Cody~Mazza-Anthony, Bogdan~Mazoure, and~Mark~Coates,~\IEEEmembership{Senior Member,~IEEE}
\thanks{This work was supported by the Department of National Defence’s
Innovation for Defence Excellence and Security (IDEaS) program. We acknowledge the support of the Natural Sciences and Engineering Research Council of Canada (NSERC), [2017-260250]. C. Mazza-Anthony and Mark Coates are with the Department of Electrical and Computer Engineering, McGill University, Montreal, QC, Canada. e-mails: cody.mazza-anthony@mail.mcgill.ca, mark.coates@mcgill.ca. B. Mazoure is with the School of Computer Science, McGill University, Montreal, QC, Canada. email: bogdan.mazoure@mail.mcgill.ca.}}

\maketitle

\begin{abstract}
We address the task of estimating sparse structured precision matrices for multivariate Gaussian random variables within a graphical model framework. We propose two novel estimators based on the Ordered Weighted $\ell_1$ (OWL) norm: 1) The Graphical OWL (GOWL) is a penalized likelihood method that applies the OWL norm to the lower triangle components of the precision matrix. 2) The column-by-column Graphical OWL (ccGOWL) estimates the precision matrix by performing OWL regularized linear regressions. Both methods can simultaneously identify groups of related edges in the graphical model and control the sparsity in the estimated precision matrix. We propose proximal descent algorithms to find the optimum for both estimators. For synthetic data where group structure is present, the ccGOWL estimator requires significantly reduced computation and achieves similar or greater accuracy than state-of-the-art estimators. Timing comparisons are presented and demonstrate the superior computational efficiency of the ccGOWL. We demonstrate the efficacy of the ccGOWL estimator on two domains---gene network analysis and econometrics.
\end{abstract}

\IEEEpeerreviewmaketitle

\section{Introduction}

The task of estimating highly modular structures based on relationships found in the data frequently arises in computational biology and finance \cite{peterman2016assessing}. Due to the large volume and high dimensionality of data in these disciplines, the speed of inference procedures and interpretability can be improved by leveraging the structured relations between inputs~\cite{stone2009modulated}. For example, it is often desirable to induce sparsity in an effort to reduce cross-pathway connections between genes when considering gene expression data. 

Gaussian graphical models (GGMs) are well-suited for modeling conditional independence through the non-zero pattern of the inverse covariance matrix. From a probabilistic point of view, the precision (inverse covariance) matrix directly encodes the conditional independence relations between its elements. The most well-known algorithm for learning sparse GGMs is the \textit{graphical lasso} \cite{friedman2008sparse}, which imposes graph sparsity through maximization of an $\ell_1$-penalized Gaussian log-likelihood with block coordinate descent. Current state-of-the-art methods \cite{hosseini2016learning, tan2015cluster,defazio2012convex,duchi2012projected} also rely on $\ell_1$ penalized structural learning. However, estimating the precision matrix column-by-column has received considerable attention since such an approach can be numerically simpler and still achieve similar performance~\cite{cai2011constrained, meinshausen2006high}. 

Related work can be organized into three classes:
1.~{\em Methods
  which learn a graph given groups a priori}: This class
includes \cite{duchi2012projected} which applies a group $\ell_1$
penalty to encourage structured sparsity but requires a set of
pre-defined hyperparameters to control the topology of the network. In other works,
prior knowledge is used to assign edge weights in order to predict
underlying group structure \cite{sun2015inferring,lee2015joint,wang2015joint,cai2016joint}.
2.~{\em Methods which first find
  groups and then learn the structure within each group}: This class
includes \cite{tan2015cluster} which proposes a two-step approach to
the problem, first applying hierarchical clustering to identify groups
and then using the graphical lasso within each
group. \cite{devijver2018block} detects groups in the covariance
matrix using thresholding and then apply graphical lasso to each
group.
3.~{\em Methods which learn both group and graph structures
  simultaneously}: \cite{defazio2012convex}  uses a
non-decreasing, concave penalty function to identify densely connected nodes in the network. \cite{hosseini2016learning} proposes
 the GRAB estimator that solves a joint optimization problem which
 alternates between estimating overlapping groups encoded into a
 Laplacian prior and learning the precision
 matrix. \cite{kumar2019unified} proposes a framework for structured
graph learning that imposes Laplacian spectral constraints.
\cite{tarzanagh2018estimation} applies a structured norm to precision
matrix estimation to identify overlapping groups. 

This paper presents two novel estimators for identifying the precision
matrices associated with structured Gaussian graphical models and makes four
major contributions. First, both estimators require only a small
amount of \textit{a priori} information and do not require any
information about group structure. More specifically, they only
require two hyperparameters (one for sparsity and one for grouping)
when using the OSCAR-like weight
generation procedure \cite{bondell2008simultaneous}. This differs from \cite{duchi2012projected},
which requires $p^2$ hyperparameters, and GRAB
\cite{hosseini2016learning}, which requires a more constrained
objective accompanied by an additional hyperparameter for each
constraint. Second, both estimators can learn network structure in a
single-step proximal descent procedure. This is an advantage over the
\textit{cluster graphical lasso} \cite{tan2015cluster}, which solves
the task with a two-step procedure by alternating between clustering
and gradient steps. Likewise, the GRAB algorithm also alternates
between learning the overlapping group prior matrix and learning the
inverse covariance. Our focus is on estimating the graphical model,
rather than identifying groups, but we can apply a Gaussian Mixture
Model (GMM) or other clustering algorithms to identify overlapping
groups from our precision matrix estimate. Third, we establish the
uniqueness of the GOWL estimator by deriving its dual
formulation. Fourth, the ccGOWL framework provides new theoretical
guarantees for grouping related entries in the precision matrix and
offers a more computationally efficient algorithm. When comparing
ccGOWL to the previously mentioned penalized likelihood methods, it is
clear that the ccGOWL estimator is more computationally attractive in the high
dimensional setting as it can be obtained one column at a time by
solving a simple linear regression that can be easily
parallelized/distributed. 

The proposed GOWL and ccGOWL estimators are based on a different
notion of groups than the algorithms discussed above. Previous methods
consider groups or blocks of {\em variables}. Structural priors then
encourage group (block) sparsity --- an edge in the graphical model (a
non-zero precision matrix entry) is modelled as more likely
if it connects variables that belong to the same group. In contrast, the GOWL estimator introduces a structural prior that encourages grouping of {\em edges} in the graphical model. The GOWL estimator favours precision matrix estimates where multiple entries have the same value, i.e., the pairs of
variables are estimated to have the same partial correlation. 
The GOWL structural prior thus models scenarios where the relationships between
multiple pairs of variables are likely to have the same strength,
with the relationships arising from a common cause or being impacted
by a common factor. As an example, consider three mining companies
which focus on different metals; we might a priori expect the
partial correlations between the stock prices and the specific
metal prices to be of the same magnitude. The ccGOWL estimator is
similar, but only encourages grouping of edges that have one variable
in common. For this case, consider an example of a single company that
devotes equal resources to mining three metals; we might model the
partial correlations between the company's stock price and the three
metals to be equal. 

\begin{figure}[h]
  \centering
  \includegraphics[width=1.0\linewidth]{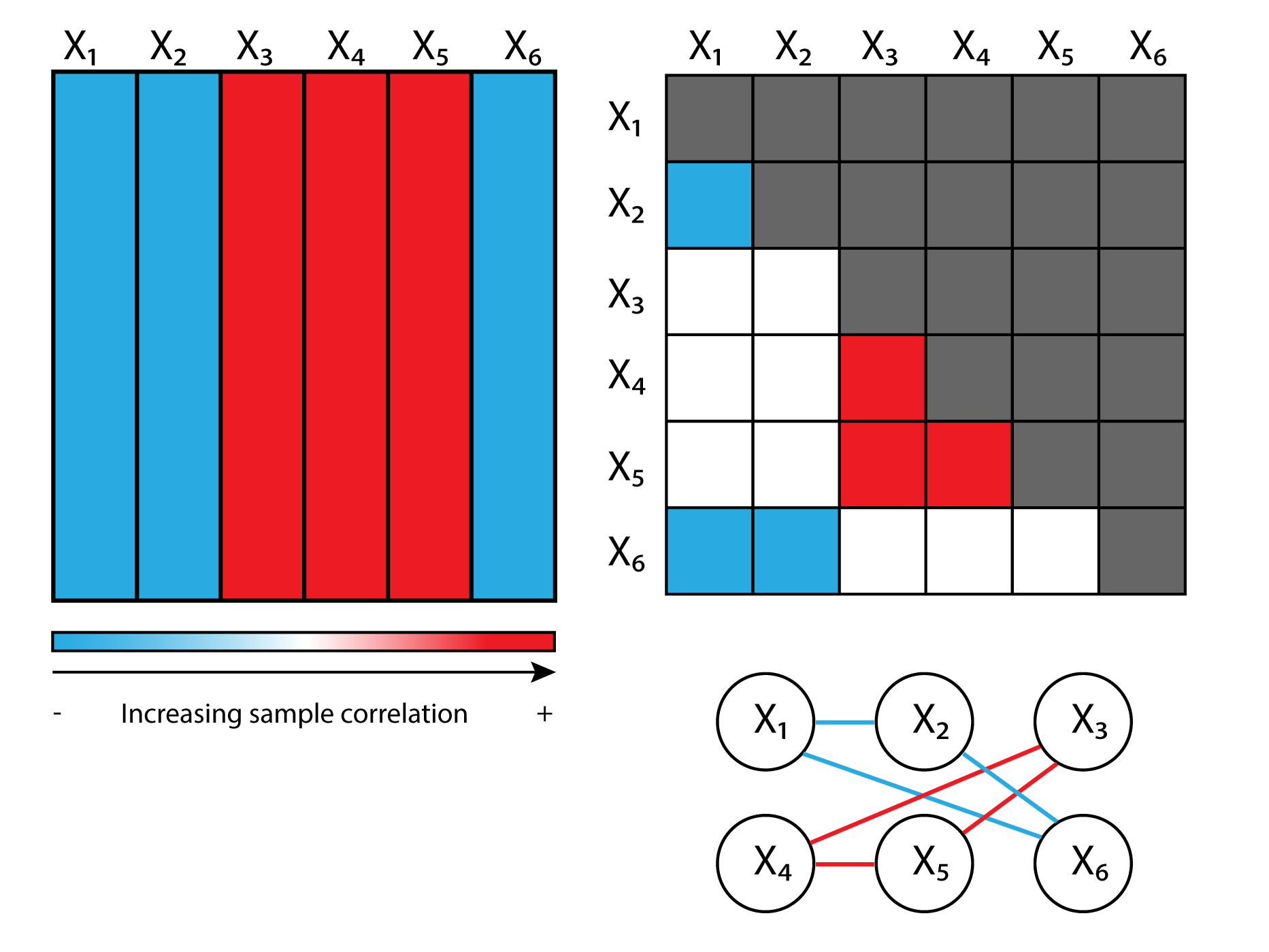}
  \caption{A design matrix, precision matrix, and network with two groups of edges that share similar sample partial correlation.}
  \label{fig:groups}
\end{figure}

Figure \ref{fig:groups} illustrates the relationship between the design matrix, precision matrix, and the resulting penalized GGM. GOWL penalizes certain groups of edges all together while ccGOWL penalizes edges for each node separately. The goal is to promote groups of related edges taking the same value while penalizing less relevant relationships between variables.

\subsection{Notation and Preliminaries.}

Throughout the paper, we highlight vectors and matrices by lowercase and uppercase boldfaced letters, respectively. For a vector $\vec{a} \in \mathbb{R}^{p}$, let $\vec{a}_{-i}$ denote a vector with its $i^{th}$ component removed. For a symmetric matrix $\mat{A} \in \mathbb{R}^{p \times p}$, $\mat{A}_{i, *}$ represents the $i^{th}$ row of $\mat{A}$ and $\mat{A}_{*, j}$ denotes the $j^{th}$ column of $\mat{A}$, $\mat{A}_{i, -j}$ denotes the $i^{th}$ row of $\mat{A}$ with its $j^{th}$ entry removed, $\mat{A}_{-i, j}$ denotes the $j^{th}$ column of $\mat{A}$ with its $i^{th}$ entry removed, and the matrix $\mat{A}_{-i, -j}$ denotes a $(p - 1) \times (p - 1)$ matrix obtained by removing the $i^{th}$ row and $j^{th}$ column. Moreover, we denote as $\text{\textbf{vechs}}(\mat{A})$ the strict column-wise vectorization of the lower triangular component of $\mat{A}$:
\begin{align*}
    \text{\textbf{vechs}} (\mat{A}) &= [\mat{A}_{2,1}, \dots, \mat{A}_{n,1}, \mat{A}_{3,2}, \dots, \mat{A}_{n,2}, \mat{A}_{n,n-1}] \,.
\end{align*}
For a vector $\vec{a} = (a_1, \dots, a_p)^T \in \mathbb{R}^p$, we use the classical definition of the $\ell_q$ norm, that is $||\vec{a}||_q = ( \sum_{i=1}^p |a_i|^q)^{1/q}$ for $1 \le q \le \infty$. The $i^{th}$ largest component in magnitude of the $p$-tuple $( |a_1|, |a_2|, \dots, |a_p| )$ is denoted $a_{[i]}$. The vector obtained by sorting (in non-increasing order) the components of $\vec{a}$ is denoted $\vec{a}_{\downarrow}=(a_{[1]},a_{[2]},...,a_{[p]})$. For matrices and vectors, we define $|\cdot|$ to be the element-wise absolute value function.

\section{Background}

\subsection{Gaussian Graphical Models (GGM)}

A GGM aims to determine the conditional independence relations of a set of random jointly Gaussian variables. Suppose $\mat{X} = [\bm{X}_1, \dots, \bm{X}_n]^T$, $\bm{X}_i \sim N(\bm{0}, \bm{\Sigma})$ is a collection of i.i.d.\ $p$-dimensional random samples. Assume that the columns of the design matrix $\mat{X}$ have been standardized to have zero mean and unit variance. Let $\bm{S}$ denote the (biased, maximum likelihood) sample covariance matrix, defined as $\bm{S} = n^{-1} \sum_{k=1}^n \bm{X}_k \bm{X}_k^T$ and let $\bm{\Theta}$ denote the precision matrix, defined as $\bm{\Theta} = \bm{\Sigma}^{-1}$. It is well understood that the sparsity pattern of $\bm{\Theta}$ encodes the pairwise partial correlation between variables. More specifically, the  $ij^\text{th}$ entry in $\bm{\Theta}$ is zero if and only if variables $i$ and $j$ are conditionally independent given the remaining components \cite{koller2009probabilistic, wytock2013sparse}. The GLASSO algorithm has been proposed to estimate the conditional independence graph embedded into $\bm{\Theta}$ through $\ell_1$ regularization of Gaussian maximum likelihood estimation:
\begin{align}
    \min_{\bm{\Theta} \succ 0} \quad - \log \text{det } \bm{\Theta} + \text{tr}(\bm{S} \bm{\Theta}) + \lambda ||\mat{\Theta}||_1 \,,
\end{align}
where $\lambda$ is a nonnegative tuning parameter that controls the level of sparsity in $\bm{\Theta}$. Under the same assumptions, \cite{meinshausen2006high} showed that $\mat{\Theta}$ can be estimated through column-by-column linear regressions. Letting $\vec{x_j} \sim N(\bm{0}, \bm{\Sigma})$ for $j = 1, \dots, n$, the conditional distribution of $x_j$ given $\vec{x}_{-j}$ satisfies:
\begin{align*}
    x_j | \vec{x}_{-j} \sim N(\bm{\beta}_j^T \vec{x}_{-j}, \sigma_j^2)\,,
\end{align*}
where $\bm{\beta}_j = ( \mat{\Sigma}_{-j, -j})^{-1} \mat{\Sigma}_{-j, j} \in \mathbb{R}^{p-1}$ and $\sigma_j^2 = \mat{\Sigma}_{j, j} - \mat{\Sigma}_{j, -j} (\mat{\Sigma}_{-j, -j}) \mat{\Sigma}_{-j, j}$. Thus, we can write the following
\begin{align*}
    x_j = \bm{\beta}_j^T \vec{x}_{-j} + \varepsilon_j\,,
\end{align*}
where $\varepsilon_j \sim N(0, \sigma_j^2)$. By the block matrix inversion formula we define:
\begin{align*}
    \mat{\Theta}_{j,j} = \sigma_j^{-2}, \quad \mathrm{and} \quad \mat{\Theta}_{-j, j} = - \sigma_j^{-2} \bm{\beta}_j \,.
\end{align*}
Thus, we can estimate the precision matrix column-by-column by regressing $x_j$ on $\vec{x}_{-j}$, and a LASSO procedure can be adopted by solving:
\begin{align}
    \hat{\bm{\beta}}_j = \argmin_{\bm{\beta}_j \in \mathbb{R}^{p-1}} || \mat{X}_{*, j} - \mat{X}_{*, -j} \bm{\beta}_j ||_2^2 + \lambda ||\bm{\beta}_j||_1 \,.
\end{align}

\subsection{OWL in the Linear Model Setting}

\begin{figure*}[htp]
\centering
\includegraphics[width=0.7\linewidth]{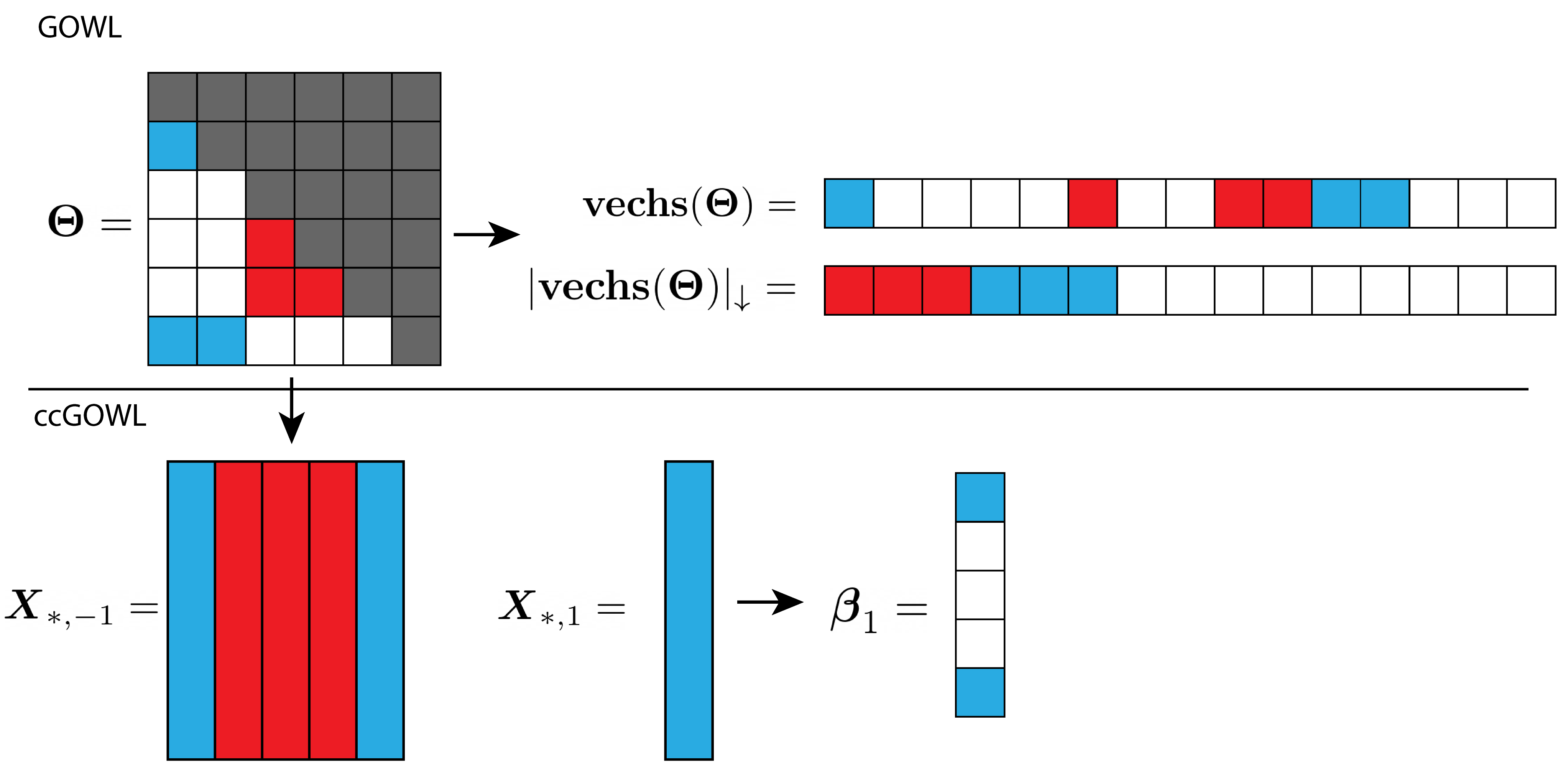}
\caption{The GOWL and ccGOWL frameworks---likelihood and column-by-column OWL penalized estimation procedures.}
\label{fig:ccgowl_gowl}
\end{figure*}

In the linear regression setting, much interest has been given to regularizers that can identify groups of highly correlated predictors. This is often referred to as \textit{structured/group sparsity}. Regularization penalties targeting such structure include the \textit{elastic net}~\cite{zou2005regularization}, the \textit{fused lasso}~\cite{tibshirani2005sparsity} and the recently proposed \textit{order weighted $\ell_1$ lasso}~\cite{bogdan2015slope, figueiredo2016ordered}. Suppose that the response vector $\bm{y} \in \mathbb{R}^n$ is generated from a linear model of the form: $\bm{y} = \bm{X} \bm{\beta} + \bm{\varepsilon}\,$ where $\bm{X} \in \mathbb{R}^{n \times p}$ is the design matrix, $\bm{\beta} \in \mathbb{R}^p$ is the vector of regression coefficients, $\bm{y} \in \mathbb{R}^n$ is the response vector, and $\bm{\varepsilon} \in \mathbb{R}^n$ is a vector of zero-centered homoscedastic random errors distributed according to $\mathcal{N}(0,\sigma^2)$. The OWL regularizer is then defined to be:
\begin{align}
    \Omega_{\text{OWL}}(\beta)=\sum_{i=1}^p \xi_i |\beta|_{[i]}\,,
\end{align}
where $\beta_{[i]}$ is the $i^\text{th}$ largest component in magnitude of $\bm{\beta} \in \mathbb{R}^p$, and $\bm{\xi} \in \mathbb{R}_{+}^p$.  OSCAR \cite{bondell2008simultaneous} is a particular case of the OWL that consists of a combination of a $\ell_1$ and a pairwise $\ell_\infty$ terms and in its original formulation is defined as 
\begin{align}
    \lambda_1 \sum_{i=1}^p |\beta_i| + \lambda_2 \sum_{i<j} \max \{ |\beta_i|, |\beta_j| \} \,.
\end{align}
The OSCAR weights can be recovered within OWL by setting $\xi_i = \lambda_1 + \lambda_2 (p - i)$ where $\lambda_1 > 0$ controls the level of sparsity and $\lambda_2 > 0$ controls the degree of grouping. In \cite{bogdan2015slope}, the OWL regularizer is referred to as \textit{sorted $\ell_1$ penalized estimation} (SLOPE).

\subsection{Proximal Methods} \label{subsec:prox}

Proximal algorithms form a class of robust methods which can help to solve composite optimization problems of the form:
\begin{align}\label{eqn:compobj}
    \min_{x \in \mathcal{X}}  \quad g(x) + h(x)\,,
\end{align}
where $\mathcal{X}$ is a Hilbert space with associated inner product $\langle \cdot, \cdot \rangle$ and  norm $|| \cdot ||$, and $g: \mathcal{X} \mapsto \mathbb{R}$ is a closed, continuously differentiable, and convex function. The function $h: \mathcal{X} \mapsto \mathbb{R}$ is a convex, not necessarily differentiable function. The \textit{proximal mapping} of a convex function $h$, denoted by $\text{prox}_h: \mathcal{X} \rightarrow \mathcal{X}$, is given by
\begin{align}
\text{prox}_h (v) &= \argmin_{x} \Big( h(x) + \frac{1}{2} ||x - v||^2_2 \Big) \,.
\end{align}
In general, this family of algorithms uses the proximal mapping to handle the non-smooth component of (\ref{eqn:compobj}) and then performs a quadratic approximation of the smooth component $g$ every iteration:
\begin{align*}
    x_{k+1} = \text{prox}_h (x_k - t_k \nabla g(x_k))\,,
\end{align*}
where $t_k$ denotes the step size at iteration $k$. \cite{rolfs2012iterative} proposed a proximal gradient method for precision matrix estimation which possesses attractive theoretical properties as well as a linear convergence rate under a suitable step size. The authors define $g,h$ as two continuous convex functions mapping the domain of covariance matrices (the positive definite cone $\mathbb{S}_{++}^p$) onto $\mathbb{R}$ and jointly optimize them using Graphical ISTA \cite{rolfs2012iterative}. G-ISTA conducts a backtracking line search for determining an optimal step size or specifies a constant step size of $\lambda_{\text{min}} (\mat{\Theta}_k)$ for problems with a small $p$. This algorithm uses the duality gap as a stopping criteria and achieves a linear rate of convergence in $O (\log \varepsilon)$ iterations to reach a tolerance of $\varepsilon$.

\section{Methodology}

\subsection{Overview of OWL Estimators}

We define the Graphical Order Weighted $\ell_1$ (GOWL) estimator to be the solution to the following constrained optimization problem:
\begin{align}\label{eqn:owl}
    \min_{\bm{\Theta} \succ 0} - \log \text{det } \bm{\Theta} + \text{tr}(\bm{S} \bm{\Theta}) + \Omega_{\text{OWL}} (\mat{\Theta})\,, 
\end{align}
where
\begin{align}
    \Omega_{\text{OWL}} (\mat{\Theta}) &= \bm{\xi}^T |\text{\textbf{vechs}}(\mat{\Theta})|_{\downarrow} = \sum_{i=1}^K \xi_i |\text{\textbf{vechs}}(\mat{\Theta})|_{[i]}\,.
\end{align}
Here $\mat{\Theta}\in \mathbb{R}^{p\times p}$, $\text{\textbf{vechs}}(\mat{\Theta})_{[i]}$ is the $i^{th}$ largest off-diagonal component in magnitude of $\bm{\Theta}$, $K = (p^2 - p)/2$, and $\xi_1 \ge \xi_2 \ge \cdots \ge \xi_p \ge 0$. The proximal mapping of $\Omega_{\text{OWL}}$ denoted by $\text{prox}_{\Omega_{\text{OWL}}}: \mathbb{R}^{p \times p} \mapsto \mathbb{R}^{p \times p}$ can be efficiently computed with $O(n \log n)$ complexity using the \textit{pool adjacent violators} (PAV) algorithm for isotonic regression outlined in \cite{zeng2014ordered}. 

In a similar way, we define the column-by-column Graphical Order Weighted $\ell_1$ (ccGOWL) estimator to be the solution to the following unconstrained optimization problem:
\begin{align}\label{eqn:ccgowl}
    \hat{\bm{\beta}}_j = \argmin_{\bm{\beta}_j \in \mathbb{R}^{p-1}} || \mat{X}_{*, j} - \mat{X}_{*, -j} \bm{\beta}_j ||_2^2 + \bm{\Omega}_{\text{OWL}} (\bm{\beta}_j)\,,
\end{align}
where $\bm{\beta}_j \in \mathbb{R}^{p-1}$, with hyperparameters defined the same way as above. We then combine the $\bm{\beta}_j$ column vectors to form $\hat{\bm{\Theta}}$. Figure \ref{fig:ccgowl_gowl} depicts the different ways that each estimator applies the OWL penalty. The GOWL estimator applies the penalty to the vectorized lower triangle of the precision matrix and the ccGOWL estimator applies the OWL penalty to the columns of the precision matrix separately.

\subsection{Uniqueness of GOWL}

In this section, we derive a dual formulation of GOWL. This
formulation plays an important role for implementing efficient
algorithms and allows us to establish uniqueness of the solution of
GOWL. Introducing the dual variable $\mat{W} \in \mathbb{R}^{p \times p}$ and letting $w_i$ denote the $i$-th entry of $\mathbf{vechs}(\mat{W})$, we can
identify the dual problem as follows (see Appendix~\ref{app:dual} for details).
\begin{align}\label{eqn:dual_owl_main}
    \max_{\mat{W}\succ 0} \quad &  \log \text{det } (\mat{S} + \mat{W}) \nonumber \\
    \text{s.t.} \quad &|w_{i}| \le \xi_{i} \,\,\,\forall i,\\
    & (\mat{S} + \mat{W}) \succ 0 \,. \nonumber
\end{align}
The following proposition, which follows from the development in Appendix~\ref{app:dual}, specifies the duality gap.
\begin{proposition}
For the optimization problem~\eqref{eqn:owl} and the dual
problem~\eqref{eqn:dual_owl_main}, the duality gap $\Delta$ given a
point $\mat{W}$ in the feasible set defined by $B_{\lambda} = \{ \mat{W} : | \mat{W}_{i} | \le \xi_i, \forall i, \mat{W}\in \mathbb{S}_{++}^p \}$ is
\begin{align}
    \Delta &= \tr(\mat{S} \mat{\Theta}) + \sum_{i=1}^K \xi_i |\mat{\Theta}|_{[i]} - p\,.
\end{align}
where $\tr$ denotes the trace operator and $\mat{W}$ denotes the dual variable in the Lagrangian formulation.
\end{proposition}
In practice, $\Delta$ is estimated using the difference between the primal and the dual and acts as a stopping criterion for the iterative procedure. As opposed to other sparse graph estimation techniques, our proposed objective function has a unique solution.
\begin{theorem}
If $\xi_{i} > 0$ for all $i$, then problem \eqref{eqn:owl} has a unique optimal point $\mat{\Theta}^*$.
\label{thm:uniqueness}
\end{theorem}
The proof is provided in Appendix~\ref{app:thm1} and involves selecting $\mat{W}$ within $\text{interior}(B_\lambda) = \{ \mat{W} : | \mat{W}_{i} | < \xi_i, \forall i, \mat{W}\in \mathbb{R}^p \}$ such that Slater's condition is satisfied.

\subsection{Algorithms}

\begin{algorithm}[htp!]
  \caption{Algorithm for GOWL}
  \label{alg:gowl}
  \begin{algorithmic}
    \Require $\mat{S}$, tolerance $\varepsilon$, $\bm{\xi}$, $t_0 > 0$, $\bm{\Theta}^0$, $c \in (0,1)$
    \While{ $\Delta > \varepsilon$  } 
    \State (1) Let $t_k$ be the largest element of $\{ c^j t_{k,0} \}$ so 
    \State that for $\mat{\Theta}^{k+1} = \text{prox}_{\Omega_{\text{OWL}}} (\mat{\Theta}^k - t_k (\mat{S} - ((\mat{\Theta}^k)^{-1}))$, 
    \State the following are satisfied:
    \begin{align*}
        \mat{\Theta}^{k+1} & \succ 0, \\
        - \log \text{det } \bm{\Theta}^{k+1} + \text{tr}(\bm{S} \bm{\Theta}^{k+1}) & \le Q_{t_k} (\mat{\Theta}^{k+1}, \mat{\Theta}^{k})\,,
    \end{align*}
    \State where
    \begin{align*}
        Q_{t_k} (\mat{\Theta}^{k+1}, \mat{\Theta}^{k}) &= - \log \text{det } \bm{\Theta} + \text{tr}(\bm{S} \bm{\Theta}) \\
        & + \text{tr}((\mat{\Theta}^{k+1} - \mat{\Theta}^{k}) (\mat{S} - (\mat{\Theta}^k)^{-1})) \\
        & + \frac{1}{2 t_k} || \mat{\Theta}^{k+1} - \mat{\Theta}^{k} ||_F^2\,.
    \end{align*}
    \State (2) Set $\bm{\Theta}_{t+1} := \text{prox}_{\text{OWL}} (\bm{\Theta}^{k} - t_k (\bm{S} - (\bm{\Theta}^k)^{-1}))$.
    \State (3) Compute the duality gap $\Delta$.
    \EndWhile
    \Ensure $\varepsilon$-optimal solution $\hat{\bm{\Theta}}$.
  \end{algorithmic}
\end{algorithm}

Our work borrows from G-ISTA discussed in Section \ref{subsec:prox} for solving the non-smooth problem defined in (\ref{eqn:owl}); the details are outlined in Algorithm~\ref{alg:gowl}. In a similar way, Algorithm~\ref{alg:ccgowl} is based on the proximal method for solving the OWL regularized linear regression proposed in \cite{zeng2014ordered} and has a convergence rate of $O(1 / k)$. Algorithm~\ref{alg:ccgowl} applies a proximal method $p$ times and combines each column to arrive at the final precision matrix estimate. Code implementing these algorithms is available at https://github.com/cmazzaanthony/ccgowl.

\begin{algorithm}[htp!]
  \caption{Algorithm for ccGOWL}
  \label{alg:ccgowl}
  \begin{algorithmic}
    \Require $\mat{X}$, tolerance $\varepsilon$, $\bm{\xi}$, $t_0  > 0$
    \For{ $j \rightarrow p$ }
    \State (1) Determine $t_k > 0$ such that the following is \State satisfied:
    \begin{align*}
        ||\mat{X}_{*, j} - \mat{X}_{*, -j} \bm{\beta}_j^{k + 1}||_2^2 \le Q_{t_k} (\bm{\beta}_j^{k}, \bm{\beta}_j^{k + 1})\,,
    \end{align*}
    \State where
    \begin{align*}
         Q_{t_k} (\bm{\beta}_j^{k}, \bm{\beta}_j^{k + 1}) &= ||\mat{X}_{*, j} - \mat{X}_{*, -j} \bm{\beta}_j^{k +
         1}||_2^2 \\
         & + 2(\bm{\beta}_j^{k} - \bm{\beta}_j^{k + 1})^T \mat{X}_{*, -j}^T (\bm{\beta}_j^{k} - \bm{\beta}_j^{k + 1}) \\
         & + \frac{t_k}{2} || \bm{\beta}_j^{k} - \bm{\beta}_j^{k + 1} ||_2^2\,.
    \end{align*}
    \State (2) Compute 
    \begin{align*}
        \nabla g (\bm{\beta}_j^{k}) := \mat{X}_{*, -j}^T (\mat{X}_{*, j} - \mat{X}_{*, -j} \bm{\beta}_j^{k}) \,.
    \end{align*}
    
    \State (3) Set $\bm{\beta}_j^{k + 1} \leftarrow \text{prox}_{\text{OWL}} ( \bm{\beta}_j^{k} - t_k \nabla g (\bm{\beta}_j^{k}) )$\,.
    
    \State (4) Exit when $||\bm{\beta}_j^{k + 1} - \bm{\beta}_j^{k}||_2 < \varepsilon$\,.
    \EndFor
    \State Combine all $\hat{\bm{\beta}}_j^{k + 1}$ to form $\hat{\bm{\Theta}}$.
  \end{algorithmic}
\end{algorithm}

\subsection{Sufficient Grouping Conditions for ccGOWL}

We can establish sufficient grouping conditions when estimating each column of $\mat{\Theta}$ by drawing on previous work for the OSCAR and OWL regularizers~\cite{figueiredo2016ordered}. The term ``grouping'' refers to components of each column estimate being equal. 
\begin{theorem} \label{proof:grouping}
Let $\hat{\beta}_k, \hat{\beta}_l > 0$ be elements in the column estimate $\hat{\bm{\beta}}_j$ of $\hat{\bm{\Theta}}_j$  estimated by ccGOWL with hyperparameter $\lambda_2$, and let $\bm{a}_k,\bm{a}_l$ be two columns of the matrix $\mat{X}_{*,-j}$ such that $\bm{1}^\top \bm{a}_k=0,\bm{1}^\top \bm{a}_l=0$ and $||\bm{a}_k||_2=1,||\bm{a}_l||_2=1$. Denote $\rho_{kl}=\bm{a}_k^\top \bm{a}_l \in [-1,1]$. Then,
\begin{equation}
    \sqrt{2-2\rho_{kl}}<\lambda_2 / ||\mat{X}_{*,j}||_2 \implies \hat{\beta}_k=\hat{\beta}_l\;.
\end{equation}
\end{theorem}
Depending on the sample correlation between covariates in $\mat{X}_{*, -j} \in \mathbb{R}^{n \times (p-1)}$ and the value of $\lambda_2$, the sufficient grouping property can be quantified according to Theorem \ref{proof:grouping}. The proof is provided in Appendix~\ref{app:thm2}.

\section{Experimental Results}

To assess the performance of the proposed algorithms, we conducted a series of
experiments: we first compared the estimates found by GOWL and ccGOWL
to those from GRAB~\cite{hosseini2016learning}  and GLASSO
\cite{friedman2008sparse} on a synthetic dataset with controlled
sparsity and grouping. Then, we compared the average weighted $F_1$
scores~\cite{manning2010introduction} obtained by all four algorithms
on random positive definite grouped matrices of varying size and group
density. Additional results for the synthetic data, including absolute
error and mean squared error metrics for prediction of precision
matrix entries, are reported in Appendix~\ref{app:ae-mse}. Hyperparameters $\lambda_1$ and $\lambda_2$ using the OSCAR weight generation defined in Section 3.3 were selected
using a 2-fold standard cross-validation procedure. Similar to
\cite{hosseini2016learning}, the regularization parameter for the GRAB
estimator was also selected using a 2-fold cross-validation procedure
(see Appendix~\ref{app:hyp} for details). All algorithms were implemented in Python and executed on an Intel i7-8700k 3.20 GHz and 32 GB of RAM.

\begin{figure}[htp!]
  \centering
  \includegraphics[width=1.0\linewidth]{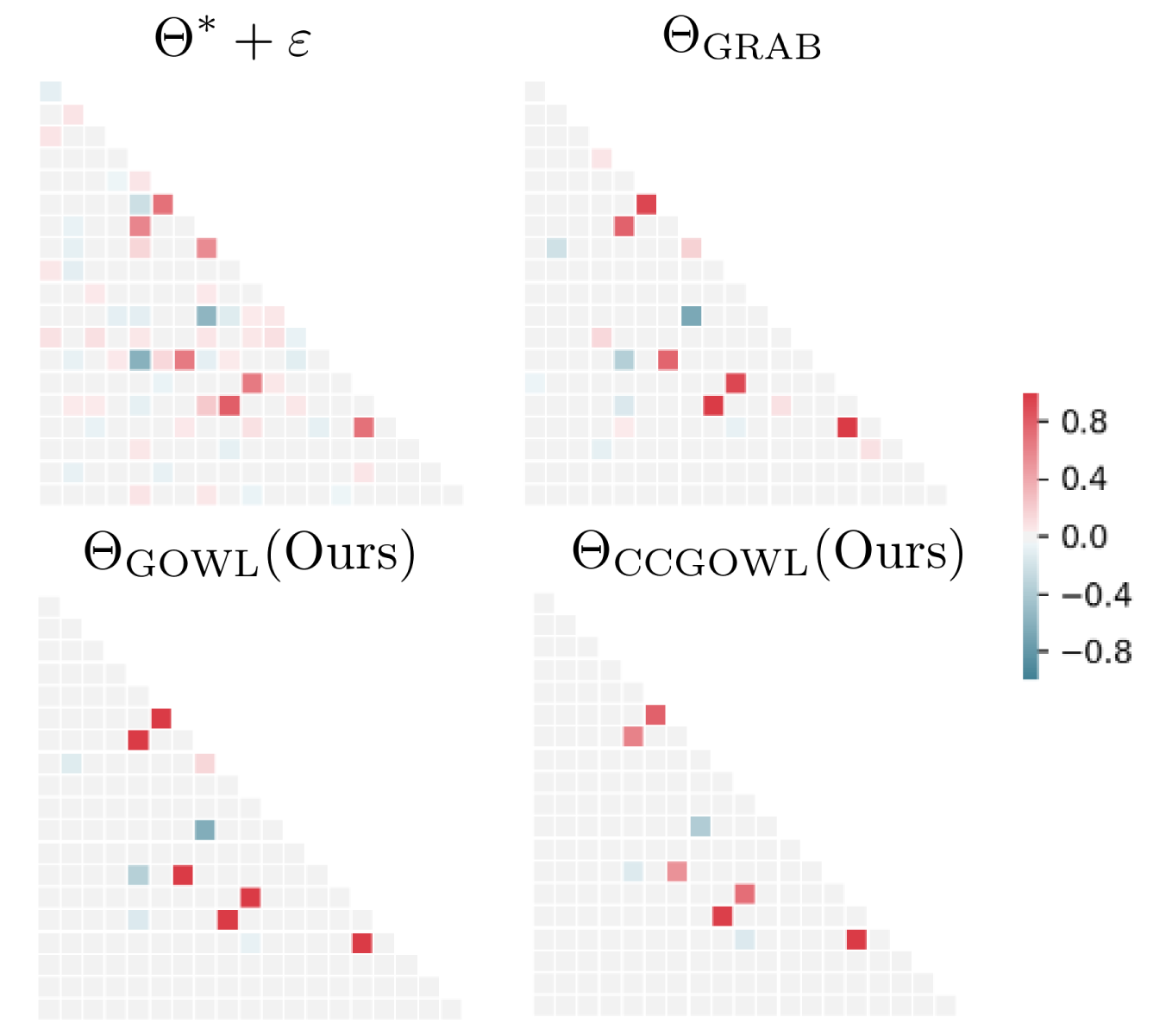}
  \caption{Synthetic data results example for methods ccGOWL/GOWL precisin matrices with $p=20$ and $\kappa = 0.1$.}
  \label{fig:heatmap}
\end{figure}

\subsection{Synthetic Data} \label{subsec:synth}
Synthetic data was generated by first creating a proportion of groups $\kappa \in \{0.1, 0.2 \}$ for a $p$-dimensional matrix. We randomly chose each group size to be between a minimum size of $0.1$ of $p$ and a maximum size of $0.4$ of $p$. Furthermore, group values were determined by uniformly sampling their mean between $(0.9, 1.0)$ and $(-0.9, -1.0)$ respectively. After setting all values of a given group to its mean, we collect the groups into the matrix $\mat{\Theta}^*$. In order to add noise to the true group values, we randomly generated a positive semi-definite matrix with entries set to zero with a fixed probability of $0.5$ and remaining values sampled between $(-0.1, 0.1)$. We then added the grouped matrix to the aforementioned noise matrix to create a $\mat{\Theta}^*+\boldsymbol{\epsilon}$ matrix. Each grouped matrix was generated 5 times and random noise matrices were added to each of the 5 grouped matrices 20 times. A dataset was then generated by drawing i.i.d.\ from a $\mathcal{N}_p (0, (\bm{\Theta}^*+\boldsymbol{\epsilon})^{-1})$ distribution. The empirical covariance matrix can be estimated after standardization of covariates. 

\begin{figure}[htp!]
  \centering
  \includegraphics[width=1.1\linewidth]{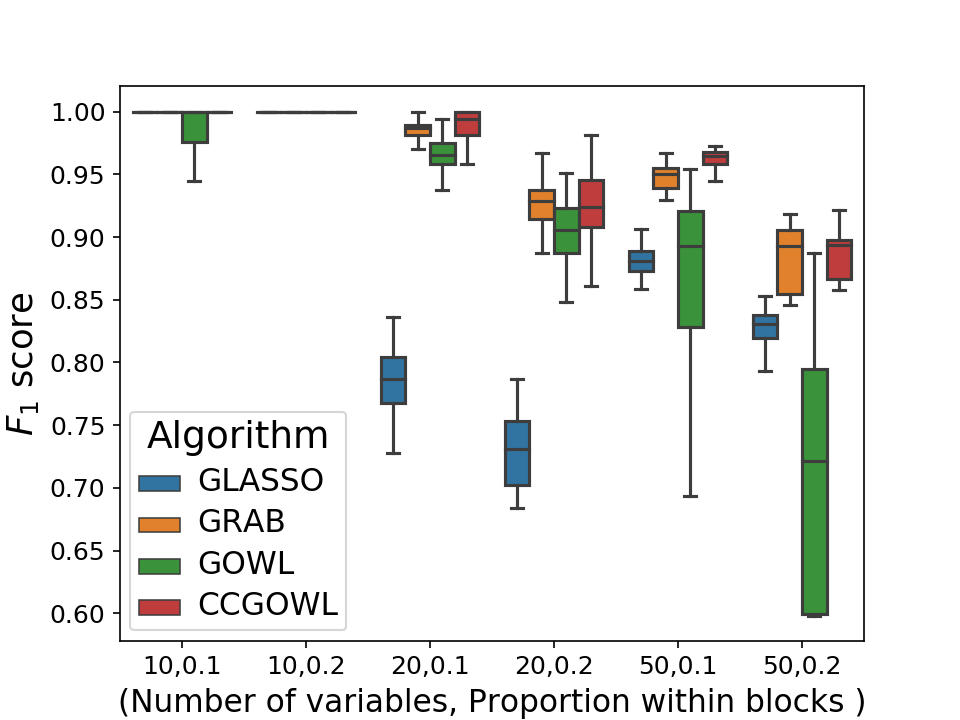}
  \caption{Quantiles of weighted $F_1$ score metric for GLASSO, GRAB, GOWL, CCGOWL.}
  \label{fig:boxplots}
\end{figure}

Figure~\ref{fig:heatmap} shows an example of a synthetic precision matrix with 2 groups with $\mat{\Theta}^*$ as the ground truth. This ground truth was then used to sample a dataset of $n=100$, from which we estimated the empirical covariance matrix and provided it as input to the algorithms. The GOWL and ccGOWL precision matrix estimates almost fully recover the 6 red entries from group 1 and the two blue entries from group two. The GLASSO was not included as a result of its poor performance in recovering the true group values. We assessed the performance of all three methods using the weighted $F_1$ classification score (harmonic mean of precision and recall) since we are interested in multi-group classification. The $F_1$ score is defined as $F_1 = 2 \cdot \frac{precision \cdot recall}{precision + recall}$; the weighted $F_1$ score is obtained by calculating the $F_1$ score for each label and then evaluating the weighted average, where the weights are proportional to the number of true instances of each group. For each value of $(p,\kappa)\in \{10,20,50\} \times \{0.1,0.2\}$, we generated 5 randomly grouped matrices using the procedure outlined in the previous section and fit GLASSO, GRAB, GOWL, ccGOWL to each of them. The estimates were then clustered using a Gaussian Mixture Model (GMM) with the same number of groups as originally set. The identification of the clusters was compared to the original group labeling using the weighted $F_1$ metric, from which we report the permutation of labels giving the highest weighted $F_1$ score. Figure~\ref{fig:boxplots} shows the distribution of the scores for each algorithm, for each class of matrices. Overall, the ccGOWL outperforms GOWL and GLASSO in terms of variance and mean of $F_1$ scores. The reason the ccGOWL outperforms GOWL could be due to the fact that with GOWL we have to generate many more parameters since we penalize the entire set of edges together as opposed to one column at a time. We observe that ccGOWL achieves similar or better performance when compared to GRAB. Mean squared error and absolute error were also measured for each estimate and are provided in Appendix~\ref{app:ae-mse}.

\begin{table}[htp!]
  \caption{Timing Comparisons in Seconds}
  \label{table:timing}
  \centering
  \resizebox{0.47\textwidth}{!}{
  \begin{tabular}{lllllll}
    \toprule
    \cmidrule(r){1-2}
    Method  & $p = 10$          & $p = 20$  & $p = 50$ & $p = 100$  & $p = 500$ & $p = 1000$ \\
    \midrule
    GLASSO  &	0.006           &	0.012   &	0.017  &	0.023   & 0.118      & 0.456    \\
    GRAB    &	0.071           &	0.096   &	0.466  &	1.243   &  51.225    & 499.225 \\
    ccGOWL  &	0.003           &	0.012   &	0.034  &	0.095   &  0.305     & 4.126 \\
    \bottomrule
  \end{tabular}}
\end{table}

\subsection{Timing Comparisons}

The GLASSO, GRAB and ccGOWL algorithms were run on synthetic datasets generated in the same way as in Section \ref{subsec:synth} with varying $p$, $n=100$, $\kappa = 0.2$ and different levels of regularization. The GRAB algorithm had a fixed duality gap of $10^{-4}$ and the ccGOWL had a fixed precision of $10^{-5}$. The GRAB algorithm was implemented in \texttt{python} and uses the \texttt{R} package \texttt{QUIC}, implemented in \texttt{C++} \cite{HSDR:NIPS2011_1249}, to find the optimum of the log-likelihood function. The ccGOWL algorithm is implemented completely in \texttt{python} and requires solving $p$ linear regressions with most of the computational complexity attributed to evaluating the proximal mapping. The GLASSO algorithm uses the \texttt{python} package \texttt{sklearn} \cite{scikit-learn}. Running times are presented in Table \ref{table:timing} and show a significant advantage of ccGOWL over GRAB. This difference for large $p$ is due to GRAB requiring matrix inversions within \texttt{QUIC} ($O (p^2)$) and applying the k-means clustering algorithm ($O(pK)$) on the rows of the block matrix $\mat{Z}$.

\subsection{Cancer Gene Expression Data}
\label{subsec:gene}

\begin{figure}
  \centering
  \begin{subfigure}[b]{0.46\textwidth}
    \includegraphics[width=0.95\linewidth]{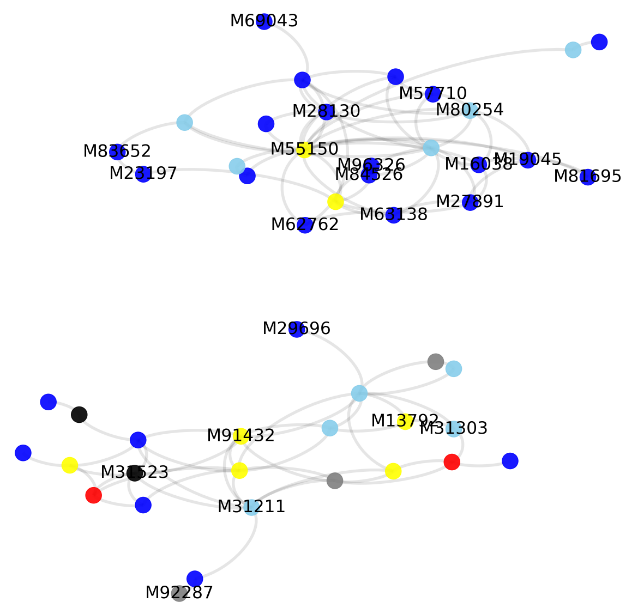}
    \caption{Network constructed by ccGOWL on gene expression data\\
    ($\lambda_1=0.3$ , $\lambda_2=0.00612821$).}
    \label{fig:gene_data_owl}
  \end{subfigure}
  \hspace{1em}
  \begin{subfigure}[b]{0.46\textwidth}  
    \includegraphics[width=0.9\linewidth]{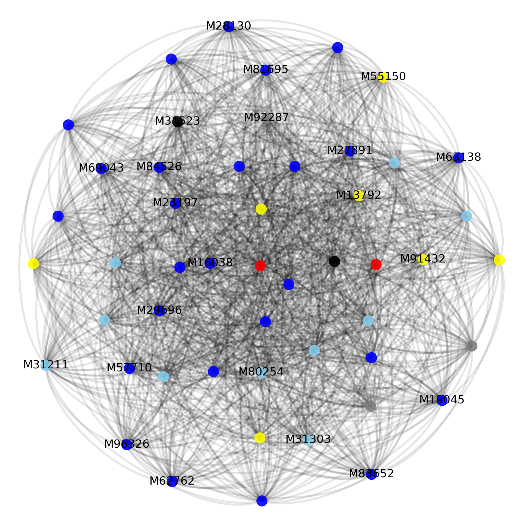}
    \caption{Network constructed by GLASSO on gene expression data
      with $\lambda=0.4$.}
      \label{fig:gene_data_glasso}
  \end{subfigure}
  \caption{The ccGOWL network estimate clearly groups genes associated with AML (top) and ALL (bottom). Each color represents a biological pathway: Signal Transduction (red), Immune System (blue), Cell Cycle (gray), Metabolism (yellow), Gene Expression (black), Uncategorized (skyblue).}
  \label{fig:gene_data}
\end{figure}

We consider a dataset that uses expression monitoring of genes using DNA microarrays in 38 patients that have been diagnosed with either acute myeloid leukemia (AML) or acute lymphoblastic leukemia (ALL). This dataset was initially investigated in \cite{golub1999molecular}. In order to allow for a model that is easier to interpret we selected the 50 most highly correlated genes associated with ALL-AML diseases, as identified in \cite{golub1999molecular}. Expression levels were standardized to have zero mean and unit variance. Of the 50 genes selected, the first 25 genes are known to be highly expressed in ALL and the remaining 25 genes are known to be highly expressed in AML. It is important to note that no single gene is consistently expressed across both AML and ALL patients. This fact illustrates the need for an estimation method that takes into account multiple genes when diagnosing patients. Edges between genes that are highly expressed in AML should appear in the same group and likewise for the ALL disease. In addition, we use the Reactome database~\cite{croft2013reactome} to identify each gene's main biological pathway.

Figure \ref{fig:gene_data_owl} illustrates that ccGOWL groups genes according to their disease. In fact, ccGOWL correctly identifies all 25 genes associated with the AML disease in one group and 24 of 25 genes associated with the ALL disease in the other group (Appendix~\ref{app:genedata}, Table~\ref{gene_expression_classes}). We compare the network identified by ccGOWL with the commonly-used baseline GLASSO method, to illustrate the importance of employing an estimator that uses grouping (Figure~\ref{fig:gene_data_glasso}). In addition, examining the connections within each group can also lead to useful insights. The AML group (top) contains highly connected genes beginning with ``M'' (blue nodes) which are associated with the neutrophil granulation process in the cell. AML is a disorder in the production of neutrophils \cite{davey1988abnormal}. Neutrophils are normal white blood cells with granules inside the cell that fight infections. AML leads to the production of immature neutrophils (referred to as blasts), leading to large infections.
\vspace{-0.1cm}
\subsection{Equities Data}
\label{subsec:stock}

\begin{figure}
\begin{subfigure}{.46\textwidth}
  \centering
  \includegraphics[width=0.9\linewidth]{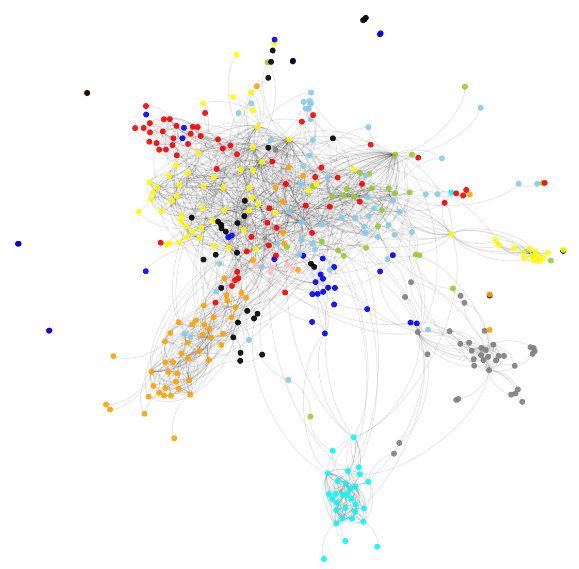}
  \caption{Network constructed by ccGOWL on equities expression data ($\lambda_1=0.4$, $\lambda_2=0.0001$).}
  \label{fig:stock_data_owl}
\end{subfigure}
\hspace{1em}
\begin{subfigure}{.46\textwidth}
  \centering
  \includegraphics[width=0.9\linewidth]{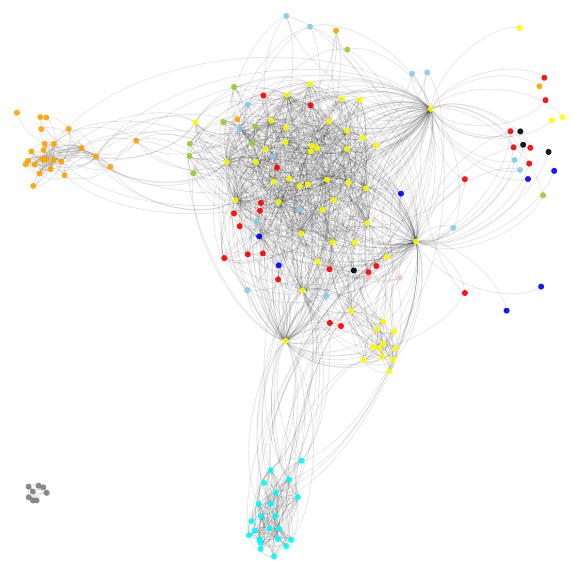}
  \caption{Network constructed by GLASSO on equities expression data with $\lambda = 0.1$.}
  \label{fig:stocks_glasso}
\end{subfigure}
\caption{Each colour represents a GICs sector: Consumer Discretionary (red), Consumer Staples (blue), Energy (gray), Financials (yellow), Health Care (black), Industrials (skyblue), Information Technology (orange), Materials (yellow-green), Telecommunications Services (pink), Utilities (cyan).}
\label{fig:stock_data}
\end{figure}

We consider the stock price dataset described in \cite{liu2012high}, which is available in the \texttt{huge} package on \texttt{CRAN} \cite{zhao2012huge}. The dataset consists of the closing prices of stocks in the S\&P 500 between January 1, 2003 and January 1, 2008. The collection of stocks can be categorized into 10 Global Industry Classification Standard (GICS) sectors \cite{gics}. Stocks that were not consistently in the S\&P 500 index or were missing too many closing prices were removed from the dataset.

The design matrix $\mat{X} \in \mathbb{R}^{1257 \times 452}$ contains the log-ratio of the price at time $t$ to the price at time $t-1$ for each of the 452 stocks for 1257 trading days. More formally we write that the $(i,j)$-th entry of $\mat{X}$ is defined as $x_{ij} = \log (S_{(i+1)j}/S_{ij})$ where $S_{ij}$ is the closing price of the $j$th stock on the $i$th day. The matrix $\mat{X}$ is then standardized so each stock has a mean of zero and unit variance. The GICS sector for each stock is known, but this information was not used when estimating the precision matrix based on the returns matrix $\mat{X}$. 
Figure \ref{fig:stock_data_owl} illustrates the ability of ccGOWL to group relationships between stocks that belong to the same sector. The identified network  is more interpretable than the network constructed by GLASSO (Figure \ref{fig:stocks_glasso}). Information Technology and Utilities largely exhibit conditional independence of stocks in other GICS sectors. On the other hand, there appears to be a conditional dependence between stocks in the Materials and Industrials sectors, probably as a result of multiple customer-supplier relationships between the companies in these sectors. Both sectors are sensitive to economic cycles and the equities offer exposure to global infrastructure replacement.

\section{Conclusion and Future Work}
We proposed the GOWL and ccGOWL estimators---two novel precision matrix estimation procedures based on the Order Weighted $\ell_1$ norm. We proved the uniqueness of the GOWL estimator and identified sufficient grouping conditions for the ccGOWL estimator. Based on empirical results on synthetic and real datasets, the ccGOWL estimator has the ability to accurately identify structure in the precision matrix in a much more computationally efficient manner than state-of-the-art estimators that achieve comparable accuracy. It is also important to mention that we believe a norm consistency result would be desirable and it could be an interesting topic of future work. This result could be achieved by following the strategy proposed in \cite{wainwright2019high} which imposes assumptions on the design matrix and deviation bound between predictors and the error term. Although penalized likelihood methods for learning structure in precision matrix estimation are widely used and perform well, it is also important to consider column-by-column estimators that accomplish similar feats while reducing computational complexity.

\appendices

\section{Dual Problem Formulation for GOWL}
\label{app:dual}
The following formulation uses the technique
from~\cite{duchi2012projected}. Let $\mat{Z} = \mat{\Theta}$ and its
associated dual variable be $\mat{W} \in \mathbb{R}^{p \times p}$,
which gives the Lagrangian:
\begin{align}
    \mathcal{L}( \mat{\Theta}, \mat{Z}, \mat{W} ) = - \log \text{det }
  \bm{\Theta} &+ \text{tr}(\bm{S} \bm{\Theta}) + \sum_{i=1}^K \xi_i
  |\text{\textbf{vechs}}(\mat{Z})|_{[i]} \nonumber \\
  &+ \text{tr}(\mat{W} ( \mat{\Theta} - \mat{Z} )),
    \label{eq:lagrange_gowl}
\end{align}
Note that the above quantity is separable into terms involving
$\mat{\Theta}$ and $\mat{Z}$, thus allowing computation of the infimum
over the auxiliary variables. We first consider the terms involving $\mat{Z}$ or more precisely, its vectorized form $\vec{z}=\text{\textbf{vechs}}(\mat{Z})$:
\begin{align*}
    g(\vec{z})=\bm{\xi}^T_{\downarrow}\vec{z}_{\downarrow}-\text{tr}(\mat{W}\mat{Z})\,.
\end{align*}
By the generalized rearrangement inequality \cite{hardy1967inequalities}, we know that for arbitrary vectors $\bm{\xi},\vec{w}$, $\bm{\xi}^T_\downarrow\vec{w}_\uparrow\leq \bm{\xi}^T\vec{w}\leq \bm{\xi}^T_\downarrow\vec{w}_\downarrow$ and hence if $\xi_i \geq |\vec{w}_i|$  for $i=1,...,K$ and $\vec{w}=\text{\textbf{vechs}}(\mat{W})$, then $\inf_{\vec{z}}g(\vec{z})=0$. Otherwise, the problem is unbounded and the infimum is attained at $g(\vec{z})=-\infty$. Therefore
\begin{align}
    \inf_{\mat{Z}} \sum_{i=1}^K \xi_i |\mat{Z}_{[i]}| - \text{tr}(\mat{W} \mat{Z}) &=
    \begin{cases}
        0 & \text{if } \xi_i \geq |\vec{w}|_i \\
        & \text{for }i=1,...,K, \\
        - \infty & \text{otherwise}\,.
    \end{cases}
    \label{eq:constraint_dual_gowl}
\end{align}
The infimum over terms involving $\mat{\Theta}$ can be computed by taking the gradient of the log determinant under the assumption that $\mat{S} + \mat{W} \succ 0$ and is attained at the point:
\begin{align}
  \inf_{\mat{\Theta}} [- \log \text{det } \bm{\Theta} &+ \text{tr}((\mat{S} +\mat{W})\mat{\Theta})]\nonumber \\
& = \log\text{det} (\mat{S} + \mat{W}) + p\,. \label{eq:dual_val}
\end{align}
Combining (\ref{eq:lagrange_gowl}) with the constraint (\ref{eq:constraint_dual_gowl}) and letting $w_i$ denote the $i$-th entry of $\mathbf{vechs}(\mat{W})$ allows us to write the dual as:
\begin{align}\label{eqn:dual_owl}
    \max_{\mat{W}\succ 0} \quad &  \log \text{det } (\mat{S} + \mat{W}) \nonumber \\
    \text{s.t.} \quad &|w_{i}| \le \xi_{i} \,\,\,\forall i,\\
    & (\mat{S} + \mat{W}) \succ 0 \,. \nonumber
\end{align}
If we define the
feasible set $B_{\lambda} = \{ \mat{W} : | \mat{W}_{i} | \le \xi_i,
\forall i, \mat{W}\in \mathbb{S}_{++}^p \}$, then for any feasible
dual point $\mat{W} \in B_{\lambda}$ the corresponding primal point
$\mat{\Theta} = (\mat{S} +
\mat{W})^{-1}$. From~\eqref{eq:constraint_dual_gowl},~\eqref{eq:dual_val}
and~\eqref{eqn:owl}, we see that the duality gap $\Delta$ is defined by:
\begin{align}
    \Delta  &=\left(-\log\text{det}(\mat{S}+\mat{W}) +\text{tr}(\mat{S} \mat{\Theta}) + \sum_{i=1}^K \xi_i
              |\mat{\Theta}|_{[i]}\right) \nonumber\\
  &\quad \quad \quad \quad \quad \quad - \left(\log \text{det}(\mat{S}+\mat{W})  + p\right) \,, \nonumber\\
  &= \text{tr}(\mat{S} \mat{\Theta}) + \sum_{i=1}^K \xi_i
              |\mat{\Theta}|_{[i]} - p
\end{align}

\section{Proof of Theorem 1}
\label{app:thm1}

The following proof uses the technique from \cite{duchi2012projected} and builds upon Slater’s condition as formulated in the following theorem:
\begin{namedtheorem}[Slater's condition]
Let a constrained optimization problem be of the form:
\begin{align}\label{eqn:nlp}
    \min_{\vec{x}\in \mathcal{X}} \quad & f(\vec{x}) \\
    s.t \quad &  g_i(\vec{x}) \le 0 \quad ( i = 1, \dots, m), \nonumber\\
    \quad & h_j(\vec{x}) = 0 \quad ( j = 1, \dots, p), \nonumber
\end{align}
We say that the Slater constraint qualification (SCQ) holds for \eqref{eqn:nlp} if there exists $\hat{\vec{x}}\in \mathcal{X}$ such that
\begin{align*}
    g_i(\hat{\vec{x}}) < 0 \quad and \quad h_j(\hat{\vec{x}}) =  0\,.
\end{align*}
Furthermore, if $\hat{\vec{x}}\in \text{interior}(\mathcal{X})$, then strong duality holds at the point $\hat{\vec{x}}$.
\end{namedtheorem}
\noindent First, assume that $c=\max_{i,j}|\mat{S}|_{ij}$ is known. In the case where $\mat{S}$ is standardized, $c=1$. The negative log-likelihood is convex in the precision matrix and is defined over the set of all positive-definite matrices. On the other hand, the OWL estimator is also convex in $\mat{\Theta}$ over the same set. Since the sum of two convex functions over the same convex set is convex, we conclude that the main objective is convex. 

Using the SCQ, we can say that the duality gap is zero and write the primal-dual optimal pair in the following way:
\begin{align*}
    \mat{\Theta}^* = (\mat{S} + \mat{W}^*)^{-1}\,.
\end{align*}
For the SCQ to hold, it remains to show that there exists a point $\mat{W}$ in the interior of the feasible set given by $\text{interior}(B_\lambda) = \{ \mat{W} : | \mat{W}_{i} | < \xi_i, \forall i, \mat{W}\in \mathbb{R}^p \}$ such that it is the solution of (\ref{eq:constraint_dual_gowl}). 

The goal is to choose a $\mat{W}$ such that $(\mat{S} + \mat{W}) \succ 0$ and ensure that the entries $|\mat{W}_i|$ are close to zero. First recall that $\mat{S}$ is a symmetric positive semi-definite matrix and since it was estimated from data, we can assume that the diagonal entries will be greater than zero with probability one. Let $\mat{A} = \text{diag} (\mat{S}) \succ 0$ since the determinant of a diagonal matrix with positive entries is positive. Consequently, by Sylvester's criterion, $\mat{A}$ is positive definite (PD). We can then write the convex combination of $\mat{S}$ and $\mat{A}$ as
\begin{align*}
    \alpha \mat{S} + (1 - \alpha) \mat{A} \succ 0\,.
\end{align*}
where $\alpha \in [0, 1)$. The above expression is itself positive definite, which we can see by taking any $\vec{x}\in \mathbb{R}^p$:
\begin{align*}
    \vec{x}^T \Big( \alpha \mat{S} + (1 - \alpha) \mat{A} \Big) \vec{x} &= \vec{x}^T (\alpha \mat{S}) \vec{x} + \vec{x}^T ((1 - \alpha) \mat{A} ) \vec{x} \\
    &= \alpha \underbrace{( \vec{x}^T \mat{S} \vec{x})}_\text{ $\ge$ 0} + (1 - \alpha) \underbrace{(\vec{x}^T  \mat{A} \vec{x})}_\text{ > 0} > 0\\
    &> 0\,.
\end{align*}
Thus, we can write
\begin{align*}
    \mat{S} + \mat{W} = \alpha \mat{S}  + (1 - \alpha) \mat{A} \succ 0,
\end{align*}
for non-negative $\alpha$ strictly smaller than 1. For a given matrix of hyperparameters $\mat{\Lambda}$, pick $\alpha>1-1/c\min_{kl} \mat{\Lambda}_{kl}$. In practice this can be achieved by setting $\tilde{\alpha}=1-1/c\min_{kl}\mat{\Lambda}_{kl}$ and putting $\alpha=\tilde{\alpha}+\varepsilon$ for some $\varepsilon>0$. Then, 
\begin{align*}
     \mat{W} &= \alpha \mat{S}  + (1 - \alpha) \mat{A} - \mat{S} \\
      &= (1 - \alpha) ( \mat{A} - \mat{S} )\,,
\end{align*}
and hence
\begin{align*}
    |\mat{W}|_{ij} &= (1 - \alpha) |\mat{S}|_{ij} \qquad (i,j = 1, \dots, p) \\
     &< 1/c\min_k \xi_k |\mat{S}_{ij}|\\
     &\leq \min_k \xi_k\\
     &\leq \xi_{ij}\,.
\end{align*}
Here we used the fact that the values in the empirical estimate of the covariance matrix $\mat{S}$ cannot be infinite. By the convexity of the primal objective and SCQ, we conclude that $\mat{W}^*$ is unique. Furthermore, since the duality gap is zero at the point $\mat{W}^*$, the uniqueness of $\mat{W}^*$ implies the uniqueness of $\mat{\Theta}^*$.
$\blacksquare$

\section{Proof of Theorem 2}
\label{app:thm2}

The following proof uses the technique from \cite{figueiredo2016ordered}. First, we define the smallest gap of consecutive elements of $\bm{\xi}$ as $\lambda_2=\min_l\{\xi_l-\xi_{l-1}:1\leq l \leq p-1\}$. Consider a pair of column estimates $\hat{\beta}_k$ and $\hat{\beta}_l$ with associated columns of the matrix $\mat{X}_{*,-j}$ denoted $a_k,a_l$ respectively, for which $||\mat{X}_{*,j}||_2||a_k-a_l||<\Delta_\lambda$. \\
\\
Let $L_2(\beta_{j})=||\mat{X}_{*,-j}\beta_{j}-\mat{X}_{*,j}||^2_2$ and let $f(\beta_{j})=L_2(\beta_{j})+\Omega_{\lambda}(\beta_{j})$. Consider the directional derivative of $f$ at a point $\hat{\beta}_{j}$ in direction $\mat{u}$. Following the argument of \cite{figueiredo2016ordered}, we see that
\begin{equation}
    f'(\hat{\mat{\beta}}_j,\mat{u})\leq ||\mat{X}_{*,j}||_2||a_k-a_l||-\lambda_2,
\end{equation}
which implies that $\hat{\beta}_j$ is not a minimizer of $f(\cdot,\mat{u})$. It follows from the contradiction argument that $\hat{\beta}_k=\hat{\beta}_l$ if $||\mat{X}_{*,j}||_2||a_k-a_l||<\lambda_2$ holds. $\blacksquare$

\section{Hyper-parameter Specification}
\label{app:hyp}

Tables~\ref{gowl_hyperparameters},~\ref{ccgowl_hyperparameters},
and~\ref{grab_hyperparameters} list the hyper-parameters chosen for
the synthetic data in Section~\ref{subsec:synth}. The hyper-parameters were chosen using
a 2-fold cross-validation procedure. For the gene dataset in Section~\ref{subsec:gene}
the hyperparameters used were $\lambda_1 = 0.3$ and
$\lambda_2 = 0.00612821$ and were chosen by using the
$\lambda_1, \lambda_2$ provided by 2-fold cross-validation and then
the sparsity hyperparameter was increased to encourage more
sparsity. For the stock dataset in Section~\ref{subsec:stock} the hyperparameters used
were $\lambda_1 = 0.2$ and $\lambda_2 = 0.0001$. The hyperparameters
were chosen arbitrarily as cross-validation required too much
computational resources.

\begin{table}
  \caption{GOWL Hyper-parameter Specification}
  \label{gowl_hyperparameters}
  \centering
  \begin{tabular}{lllll}
    \toprule
    \cmidrule(r){1-2}
    $p$      & Block \% & $\lambda_1$     & $\lambda_2$     & Range Considered \\
    \midrule
    10      & 10        & 0.03684211    & 0.01052632        & (0, 0.1) \\ 
    10      & 20        & 0.06842105    & 0.01052632        & (0, 0.1) \\
    20      & 10        & 0.06551724    & 0.00344828        & (0, 0.1) \\
    20      & 20        & 0.02413793    & 0.00344828        & (0, 0.1) \\
    50      & 10        & 0.008         & 0.00010           & (0, 0.1) \\
    50      & 20        & 0.006         & 0.00009           & (0, 0.1) \\
    \bottomrule
  \end{tabular}
\end{table}

\begin{table}
  \caption{ccGOWL Hyper-parameter Specification}
  \label{ccgowl_hyperparameters}
  \centering
  \begin{tabular}{lllll}
    \toprule
    \cmidrule(r){1-2}
    $p$      & Block \% & $\lambda_1$     & $\lambda_2$     & Range Considered \\
    \midrule
    10      & 10        & 0.10526316    & 0.05263158        & (0, 0.1) \\ 
    10      & 20        & 0.10526316    & 0.05263158        & (0, 0.1) \\
    20      & 10        & 0.23684211    & 0.00793103        & (0, 0.1) \\
    20      & 20        & 0.23684211    & 0.00793103        & (0, 0.1) \\
    50      & 10        & 0.1           & 0.00512821        & (0, 0.1) \\
    50      & 20        & 0.1           & 0.00512821        & (0, 0.1) \\
    \bottomrule
  \end{tabular}
\end{table}

\begin{table}[h!]
  \caption{GRAB Hyper-parameter Specification}
  \label{grab_hyperparameters}
  \centering
  \begin{tabular}{lllll}
    \toprule
    \cmidrule(r){1-2}
    $p$      & Block \% & $\lambda$ & Range Considered \\
    \midrule
    10      & 10        & 0.1       & (0, 1.0) \\ 
    10      & 20        & 0.1       & (0, 1.0) \\
    20      & 10        & 0.7       & (0, 1.0) \\
    20      & 20        & 0.5       & (0, 1.0) \\
    50      & 10        & 0.5       & (0, 1.0) \\
    50      & 20        & 0.4       & (0, 1.0) \\
    \bottomrule
  \end{tabular}
\end{table}

\section{Synthetic data: additional experimental results}
\label{app:ae-mse}

\begin{figure}[h]
\begin{minipage}{.5\textwidth}
  \centering
  \includegraphics[width=1.0\linewidth]{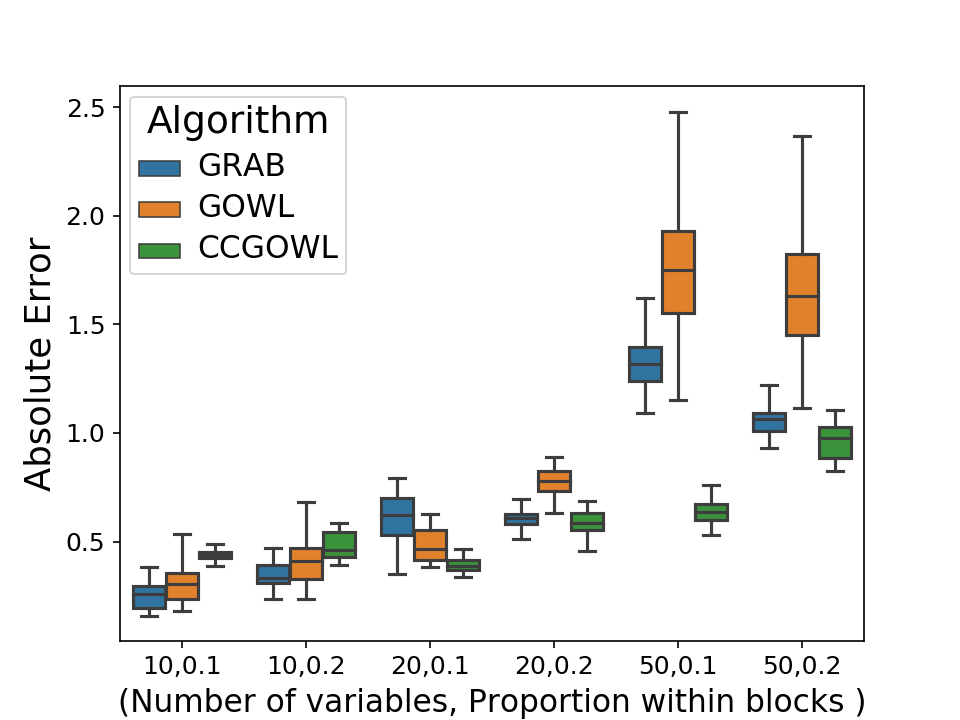}
  \caption{Absolute Error Values for GRAB, GOWL, and ccGOWL for the
    synthetic data in Section~\ref{subsec:synth}.}
  \label{fig:ase}
\end{minipage}
\hspace{1em}%
\begin{minipage}{.5\textwidth}
  \centering
  \includegraphics[width=1.0\linewidth]{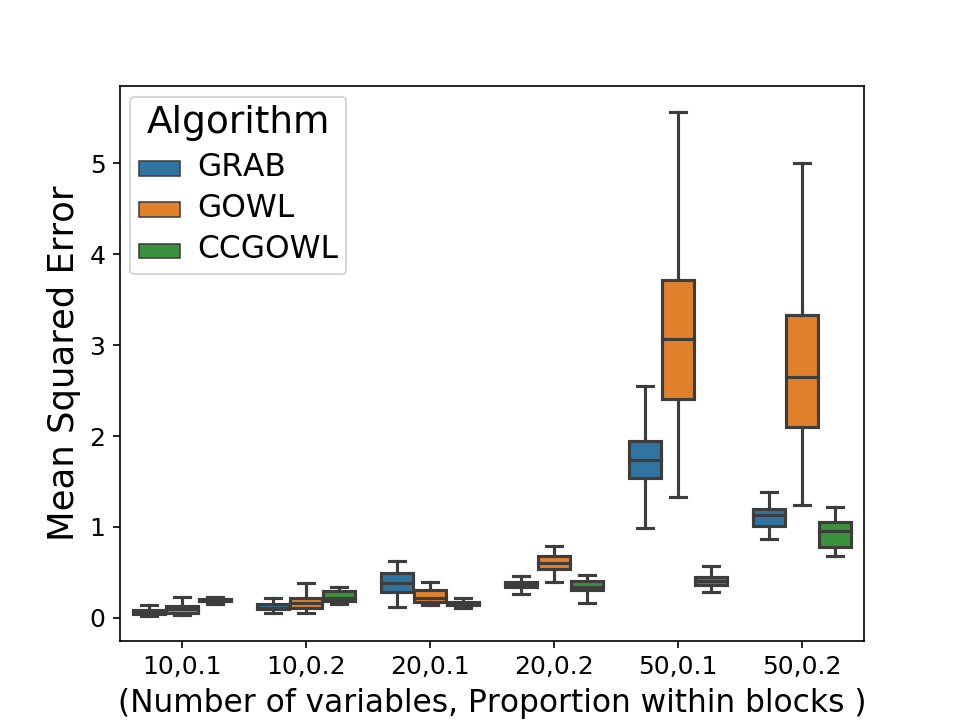}
  \caption{Mean Squared Error Values for GRAB, GOWL, and ccGOWL for the
    synthetic data in Section~\ref{subsec:synth}.}
  \label{fig:mse}
\end{minipage}
\label{fig:fig}
\end{figure}

Figures~\ref{fig:ase} and~\ref{fig:mse} show the absolute error and mean squared error for estimation of the precision matrix entries for the synthetic data experiments in Section~\ref{subsec:synth}. The ccGOWL estimator has a lower mean squared error and absolute error for $p=20$ and $p=50$ and has slightly higher errors for $p=10$. For larger values of $p$, the performance discrepancy between ccGOWL and GRAB is more noticeable for these metrics than for the weighted $F_1$ scores.

Table ~\ref{tab:f1_mse_additional} and ~\ref{tab:Sens_Spec} provide additional results, comparing the $F_1$, MSE, sensitivity and specificity of the three methods while varying parameters $p$, $\kappa$, and $n$. These Tables illustrate that the ccGOWL method has lower mean square error in most cases even for a wider range of parameters. Likewise, the specificity and sensitivity values are also larger in most cases.

    \begin{table}[htp!]
        \centering
        
        \begin{tabular}{lllrrrrrr}
        \toprule
        & & &  \multicolumn{2}{c}{GRAB} &  \multicolumn{2}{c}{GOWL} &  \multicolumn{2}{c}{CCGOWL} \\
         p &  $\kappa$ &     n & $F_1$ & MSE & $F_1$ & MSE & $F_1$ & MSE\\
         \midrule
         15 &       0.1 &  1000 &        1.00 &      2.01 &        1.00 &      5.96 &          1.00 &        2.43 \\
         15 &       0.1 &  2000 &        1.00 &      1.90 &        1.00 &      6.99 &   1.00 &        1.57 \\
         15 &       0.3 &  1000 &        0.42 &      3.06 &        0.48 &      6.65 &          0.61 &        1.51 \\ 
         15 &       0.3 &  2000 &        0.50 &      2.39 &        0.39 &      7.42 &          0.60 &        1.62 \\ 
         25 &       0.1 &  1000 &        0.71 &      4.55 &        0.89 &      8.13 &          0.87 &        0.80 \\
         25 &       0.1 &  2000 &        0.73 &      4.99 &        0.69 &      6.23 &          0.71 &        1.78 \\
         25 &       0.3 &  1000 &        0.30 &      1.04 &        0.14 &      6.20 &          0.26 &        1.23 \\
         25 &       0.3 &  2000 &        0.23 &      1.05 &        0.20 &      6.25 &          0.30 &        3.10 \\
         \bottomrule
       \end{tabular}
        
        \caption{Additional Simulations: $F_1$ and MSE scores over 10 replications.}
        \label{tab:f1_mse_additional}
    \end{table}
    
    \begin{table}[htp!]
        \centering
        \begin{tabular}{lllrrrrrr}
        \toprule
        & & &  \multicolumn{2}{c}{GRAB} &  \multicolumn{2}{c}{GOWL} &  \multicolumn{2}{c}{CCGOWL} \\
         p &  $\kappa$ &     n & Sens. & Spec. & Sens. & Spec. & Sens. & Spec.\\
         \midrule
         15 &       0.1 &  1000 &              1.00 &              1.00 &               1.00 &              1.00 &                1.00 &                1.00 \\
         15 &       0.1 &  2000 &              1.00 &              1.00 &               1.00 &              1.00 &                1.00 &                1.00 \\
         15 &       0.3 &  1000 &              0.30 &              0.34 &               0.20 &              0.47 &                0.33 &                0.90 \\
         15 &       0.3 &  2000 & 0.30 &              0.47 &               0.10 &              0.38 &                0.55 &                0.66 \\
         25 &       0.1 &  1000 &              0.59 &              0.98 &               1.00 &              0.97 &                0.86 &                0.99 \\
         25 &       0.1 &  2000 &              0.73 &              0.97 &               0.60 &              0.97 &                0.50 &                0.99 \\
         25 &       0.3 &  1000 &              0.00 &              0.33 &               0.00 &              0.20 &                0.30 &                0.20 \\
         25 &       0.3 &  2000 &              0.20 &              0.30 &               0.10 &              0.52 &                0.30 &                0.35 \\
         \bottomrule
       \end{tabular}
        
        \caption{Additional Simulations: Sensitivity and specificity scores over 10 replications.}
        \label{tab:Sens_Spec}
    \end{table}

\section{Gene Expression Data}
\label{app:genedata}
Table \ref{gene_expression_classes} shows which genes are highly expressed in individuals with either the ALL or AML disease as demonstrated in \cite{golub1999molecular}. 

\begin{table}[h!]
  \caption{Gene Classification}
  \label{gene_expression_classes}
  \centering
  \begin{tabular}{ll}
    \toprule
    \cmidrule(r){1-2}
    ALL         & AML \\
    \midrule
    U22376      & M55150         \\ 
    X59417      & X95735         \\
    U05259      & U50136        \\
    M92287      & M16038         \\ 
    M31211      & U82759         \\
    X74262      & M23197        \\
    D26156      & M84526         \\ 
    S50223      & Y12670         \\
    M31523      & M27891        \\
    L47738      & X17042         \\ 
    U32944      & Y00787         \\
    Z15115      & M96326        \\
    X15949      & U46751         \\ 
    X63469      & M80254         \\
    M91432      & L08246        \\
    U29175      & M62762         \\
    Z69881      & M28130        \\
    U20998      & M63138         \\
    D38073      & M57710        \\
    U26266      & M69043         \\
    M31303      & M81695        \\
    Y08612      & X85116         \\
    U35451      & M19045        \\
    M29696      & M83652        \\
    M13792      & X04085        \\
    \bottomrule
  \end{tabular}
\end{table}



\newpage
\bibliographystyle{IEEEtran}

\begin{thebibliography}{10}
\providecommand{\url}[1]{#1}
\csname url@samestyle\endcsname
\providecommand{\newblock}{\relax}
\providecommand{\bibinfo}[2]{#2}
\providecommand{\BIBentrySTDinterwordspacing}{\spaceskip=0pt\relax}
\providecommand{\BIBentryALTinterwordstretchfactor}{4}
\providecommand{\BIBentryALTinterwordspacing}{\spaceskip=\fontdimen2\font plus
\BIBentryALTinterwordstretchfactor\fontdimen3\font minus
  \fontdimen4\font\relax}
\providecommand{\BIBforeignlanguage}[2]{{%
\expandafter\ifx\csname l@#1\endcsname\relax
\typeout{** WARNING: IEEEtran.bst: No hyphenation pattern has been}%
\typeout{** loaded for the language `#1'. Using the pattern for}%
\typeout{** the default language instead.}%
\else
\language=\csname l@#1\endcsname
\fi
#2}}
\providecommand{\BIBdecl}{\relax}
\BIBdecl

\bibitem{peterman2016assessing}
W.~E. Peterman, B.~H. Ousterhout, T.~L. Anderson, D.~L. Drake, R.~D. Semlitsch,
  and L.~S. Eggert, ``Assessing modularity in genetic networks to manage
  spatially structured metapopulations,'' \emph{Ecosphere}, vol.~7, no.~2,
  2016.

\bibitem{stone2009modulated}
E.~A. Stone and J.~F. Ayroles, ``Modulated modularity clustering as an
  exploratory tool for functional genomic inference,'' \emph{PLoS Genetics},
  vol.~5, no.~5, p. e1000479, 2009.

\bibitem{friedman2008sparse}
J.~Friedman, T.~Hastie, and R.~Tibshirani, ``Sparse inverse covariance
  estimation with the graphical lasso,'' \emph{Biostatistics}, vol.~9, no.~3,
  pp. 432--441, 2008.

\bibitem{hosseini2016learning}
M.~J. Hosseini and S.-I. Lee, ``Learning sparse {G}aussian graphical models
  with overlapping blocks,'' in \emph{Advances in Neural Information Processing
  Systems}, 2016, pp. 3808--3816.

\bibitem{tan2015cluster}
K.~M. Tan, D.~Witten, and A.~Shojaie, ``The cluster graphical lasso for
  improved estimation of {G}aussian graphical models,'' \emph{Computational
  Statistics \& Data Analysis}, vol.~85, pp. 23--36, 2015.

\bibitem{defazio2012convex}
A.~Defazio and T.~S. Caetano, ``A convex formulation for learning scale-free
  networks via submodular relaxation,'' in \emph{Advances in Neural Information
  Processing Systems}, 2012, pp. 1250--1258.

\bibitem{duchi2012projected}
J.~Duchi, S.~Gould, and D.~Koller, ``Projected subgradient methods for learning
  sparse {G}aussians,'' in \emph{Proc. Int. Conf. Artificial Intelligence and
  Statistics}, 2008, pp. 153--160.

\bibitem{cai2011constrained}
T.~Cai, W.~Liu, and X.~Luo, ``A constrained $\ell_1$ minimization approach to
  sparse precision matrix estimation,'' \emph{Journal of the American
  Statistical Association}, vol. 106, no. 494, pp. 594--607, 2011.

\bibitem{meinshausen2006high}
N.~Meinshausen, P.~B{\"u}hlmann \emph{et~al.}, ``High-dimensional graphs and
  variable selection with the lasso,'' \emph{The Annals of Statistics},
  vol.~34, no.~3, pp. 1436--1462, 2006.

\bibitem{sun2015inferring}
S.~Sun, H.~Wang, and J.~Xu, ``Inferring block structure of graphical models in
  exponential families,'' in \emph{Proc. Int. Conf. Artificial Intelligence and
  Statistics}, 2015, pp. 939--947.

\bibitem{lee2015joint}
W.~Lee and Y.~Liu, ``Joint estimation of multiple precision matrices with
  common structures,'' \emph{The {J}ournal of {M}achine {L}earning {R}esearch},
  vol.~16, no.~1, pp. 1035--1062, 2015.

\bibitem{wang2015joint}
J.~Wang, ``Joint estimation of sparse multivariate regression and conditional
  graphical models,'' \emph{{S}tatistica {S}inica}, pp. 831--851, 2015.

\bibitem{cai2016joint}
T.~T. Cai, H.~Li, W.~Liu, and J.~Xie, ``Joint estimation of multiple
  high-dimensional precision matrices,'' \emph{{S}tatistica {S}inica}, vol.~26,
  no.~2, p. 445, 2016.

\bibitem{devijver2018block}
E.~Devijver and M.~Gallopin, ``Block-diagonal covariance selection for
  high-dimensional {G}aussian graphical models,'' \emph{Journal of the American
  Statistical Association}, vol. 113, no. 521, pp. 306--314, 2018.

\bibitem{kumar2019unified}
S.~Kumar, J.~Ying, J.~V. d.~M. Cardoso, and D.~Palomar, ``A unified framework
  for structured graph learning via spectral constraints,'' \emph{arXiv
  preprint arXiv:1904.09792}, 2019.

\bibitem{tarzanagh2018estimation}
D.~A. Tarzanagh and G.~Michailidis, ``Estimation of graphical models through
  structured norm minimization,'' \emph{Journal of {M}achine {L}earning
  {R}esearch}, vol.~18, no.~1, 2018.

\bibitem{bondell2008simultaneous}
H.~D. Bondell and B.~J. Reich, ``Simultaneous regression shrinkage, variable
  selection, and supervised clustering of predictors with {OSCAR},''
  \emph{Biometrics}, vol.~64, no.~1, pp. 115--123, 2008.

\bibitem{koller2009probabilistic}
D.~Koller, N.~Friedman, and F.~Bach, \emph{Probabilistic graphical models:
  principles and techniques}.\hskip 1em plus 0.5em minus 0.4em\relax MIT Press,
  2009.

\bibitem{wytock2013sparse}
M.~Wytock and Z.~Kolter, ``Sparse {G}aussian conditional random fields:
  Algorithms, theory, and application to energy forecasting,'' in \emph{Proc.
  Int. Conf. Machine Learning}, 2013, pp. 1265--1273.

\bibitem{zou2005regularization}
H.~Zou and T.~Hastie, ``Regularization and variable selection via the elastic
  net,'' \emph{{J}ournal of the {R}oyal {S}tatistical {S}ociety: {S}eries {B}},
  vol.~67, no.~2, pp. 301--320, 2005.

\bibitem{tibshirani2005sparsity}
R.~Tibshirani, M.~Saunders, S.~Rosset, J.~Zhu, and K.~Knight, ``Sparsity and
  smoothness via the fused lasso,'' \emph{{J}ournal of the {R}oyal
  {S}tatistical {S}ociety: {S}eries {B}}, vol.~67, no.~1, pp. 91--108, 2005.

\bibitem{bogdan2015slope}
M.~Bogdan, E.~Van Den~Berg, C.~Sabatti, W.~Su, and E.~J. Cand{\`e}s,
  ``{SLOPE}—adaptive variable selection via convex optimization,'' \emph{The
  Annals of Applied Statistics}, vol.~9, no.~3, p. 1103, 2015.

\bibitem{figueiredo2016ordered}
M.~Figueiredo and R.~Nowak, ``Ordered weighted $\ell_1$ regularized regression
  with strongly correlated covariates: Theoretical aspects,'' in \emph{Proc.
  Int. Conf. Artificial Intelligence and Statistics}, 2016, pp. 930--938.

\bibitem{rolfs2012iterative}
B.~Rolfs, B.~Rajaratnam, D.~Guillot, I.~Wong, and A.~Maleki, ``Iterative
  thresholding algorithm for sparse inverse covariance estimation,'' in
  \emph{Advances in Neural Information Processing Systems}, 2012, pp.
  1574--1582.

\bibitem{zeng2014ordered}
X.~Zeng and M.~A. Figueiredo, ``The ordered weighted $\ell_1$ norm: Atomic
  formulation, projections, and algorithms,'' \emph{arXiv preprint
  arXiv:1409.4271}, 2014.

\bibitem{manning2010introduction}
C.~Manning, P.~Raghavan, and H.~Sch{\"u}tze, ``Introduction to information
  retrieval,'' \emph{Natural Language Engineering}, vol.~16, no.~1, pp.
  100--103, 2010.

\bibitem{HSDR:NIPS2011_1249}
C.-J. Hsieh, M.~A. Sustik, I.~S. Dhillon, and P.~K. Ravikumar, ``Sparse inverse
  covariance matrix estimation using quadratic approximation,'' in
  \emph{Advances in Neural Information Processing Systems}, 2011, pp.
  2330--2338.

\bibitem{scikit-learn}
F.~Pedregosa, G.~Varoquaux, A.~Gramfort, V.~Michel, B.~Thirion, O.~Grisel,
  M.~Blondel, P.~Prettenhofer, R.~Weiss, V.~Dubourg, J.~Vanderplas, A.~Passos,
  D.~Cournapeau, M.~Brucher, M.~Perrot, and E.~Duchesnay, ``Scikit-learn:
  Machine learning in {P}ython,'' \emph{Journal of Machine Learning Research},
  vol.~12, pp. 2825--2830, 2011.

\bibitem{golub1999molecular}
T.~R. Golub, D.~K. Slonim, P.~Tamayo, C.~Huard, M.~Gaasenbeek, J.~P. Mesirov,
  H.~Coller, M.~L. Loh, J.~R. Downing, M.~A. Caligiuri \emph{et~al.},
  ``Molecular classification of cancer: class discovery and class prediction by
  gene expression monitoring,'' \emph{Science}, vol. 286, no. 5439, pp.
  531--537, 1999.

\bibitem{croft2013reactome}
D.~Croft, A.~F. Mundo, R.~Haw, M.~Milacic, J.~Weiser, G.~Wu, M.~Caudy,
  P.~Garapati, M.~Gillespie, M.~R. Kamdar \emph{et~al.}, ``The reactome pathway
  knowledgebase,'' \emph{Nucleic Acids Research}, vol.~42, no.~D1, pp.
  D472--D477, 2013.

\bibitem{davey1988abnormal}
F.~R. Davey, W.~N. Erber, K.~C. Gatter, and D.~Y. Mason, ``Abnormal neutrophils
  in acute myeloid leukemia and myelodysplastic syndrome,'' \emph{Human
  Pathology}, vol.~19, no.~4, pp. 454--459, 1988.

\bibitem{liu2012high}
H.~Liu, F.~Han, M.~Yuan, J.~Lafferty, L.~Wasserman \emph{et~al.},
  ``High-dimensional semiparametric {G}aussian copula graphical models,''
  \emph{The Annals of Statistics}, vol.~40, no.~4, pp. 2293--2326, 2012.

\bibitem{zhao2012huge}
T.~Zhao, H.~Liu, K.~Roeder, J.~Lafferty, and L.~Wasserman, ``The huge package
  for high-dimensional undirected graph estimation in {R},'' \emph{{J}ournal of
  {M}achine {L}earning {R}esearch}, vol.~13, no. Apr, pp. 1059--1062, 2012.

\bibitem{gics}
\BIBentryALTinterwordspacing
MSCI, ``The {G}lobal {I}ndustry {C}lassification {S}tandard ({GICS}),'' 2019.
  [Online]. Available: \url{https://www.msci.com/gics,}
\BIBentrySTDinterwordspacing

\bibitem{wainwright2019high}
M.~J. Wainwright, \emph{High-dimensional statistics: A non-asymptotic
  viewpoint}.\hskip 1em plus 0.5em minus 0.4em\relax Cambridge University
  Press, 2019, vol.~48.

\bibitem{hardy1967inequalities}
G.~Hardy, J.~Littlewood, and G.~P{\'o}lya, ``Inequalities. {C}ambridge
  {M}athematical {L}ibrary {S}eries,'' 1967.

\end{thebibliography}
\end{document}